\def\year{2020}\relax
\documentclass[letterpaper]{article} 
\usepackage{aaai20}  
\usepackage{times}  
\usepackage{helvet} 
\usepackage{courier}  
\usepackage[hyphens]{url}  
\usepackage{graphicx} 
\usepackage{graphicx}  

\usepackage{subcaption}
\usepackage{booktabs}
\usepackage{multirow}
\usepackage{amsmath}

\newcommand{\Eg}{\textit{E}.\textit{g}.}

\title{DWM: A Decomposable Winograd Method for Convolution Acceleration}

\author{
Di Huang,\textsuperscript{\rm 1,2,3}
Xishan Zhang,\textsuperscript{\rm 1}\thanks{Xishan Zhang (zhangxishan@ict.ac.cn) is the corresponding author.}
Rui Zhang,\textsuperscript{\rm 1}
Tian Zhi,\textsuperscript{\rm 1}
Deyuan He,\textsuperscript{\rm 1,2,3}
Jiaming Guo,\textsuperscript{\rm 1,2,3}\\
\Large \textbf{
Chang Liu,\textsuperscript{\rm 1,2,3}
Qi Guo,\textsuperscript{\rm 1}
Zidong Du,\textsuperscript{\rm 1}
Shaoli Liu,\textsuperscript{\rm 3}
Tianshi Chen,\textsuperscript{\rm 3}
Yunji Chen\textsuperscript{\rm 1,2,4,5,6}}\\
\textsuperscript{\rm 1}SKL of Computer Architecture, Institute of Computing Technology, CAS\\
\textsuperscript{\rm 2}University of Chinese Academy of Sciences,
\textsuperscript{\rm 3}Cambricon Tech. Ltd\\
\textsuperscript{\rm 4}Institute of Brain-Intelligence Technology, Zhangjiang Laboratory\\
\textsuperscript{\rm 5}Shanghai Research Center for Brian Science and Brain-Inspired Instelligence\\
\textsuperscript{\rm 6}CAS Center for Excellence in Brain Science and Intelligence Technology\\
\{huangdi18s, zhangxishan, zhangrui, zhitian, hedeyuan18s, guojiaming18s, liuchang18s, guoqi, duzidong, cyj\}@ict.ac.cn\\
\{liushaoli, tchen\}@cambricon.com
}
\begin{document}
\maketitle
\begin{abstract}
Winograd's minimal filtering algorithm has been widely used in Convolutional Neural Networks (CNNs) to reduce the number of multiplications for faster processing. 
However, it is only effective on convolutions with kernel size as 3x3 and stride as 1, because it suffers from significantly increased FLOPs and numerical accuracy problem for kernel size larger than 3x3 and fails on convolution with stride larger than 1.
In this paper, we propose a novel Decomposable Winograd Method (DWM), which breaks through the limitation of original Winograd's minimal filtering algorithm to a wide and general convolutions. 
DWM decomposes kernels with large size or large stride to several small kernels with stride as 1 for further applying Winograd method, so that DWM can reduce the number of multiplications while keeping the numerical accuracy.
It enables the fast exploring of larger kernel size and larger stride value in CNNs for high performance and accuracy and even the potential for new CNNs.
Comparing against the original Winograd, the proposed DWM is able to support all kinds of convolutions with a speedup of $\sim$2, without affecting the numerical accuracy.
\end{abstract}
\section{Introduction}
Deep Convolutional Neural Networks (CNNs) have shown excellent performance on many machine learning tasks but has been plagued by the huge amount of computations for a long time.
Recently, CNNs are increasingly larger to achieve better accuracy but also increase the amount of computations. \Eg, AlexNet~\cite{krizhevsky2012imagenet}, has $7\times10^8$ multiplications, VGG-16~\cite{simonyan2014very} contains $1.5\times10^{10}$ multiplications, and SENet-154~\cite{hu2018squeeze} contains $2.1\times10^{10}$ multiplications.
The massive amount of computations blocks the application of CNNs.
Therefore, reducing the multiplications in CNNs is essential and meaningful.

\begin{figure}[thb!]
    \centering
    \begin{subfigure}{\columnwidth}
        \centering
        \includegraphics[width=0.95\columnwidth]{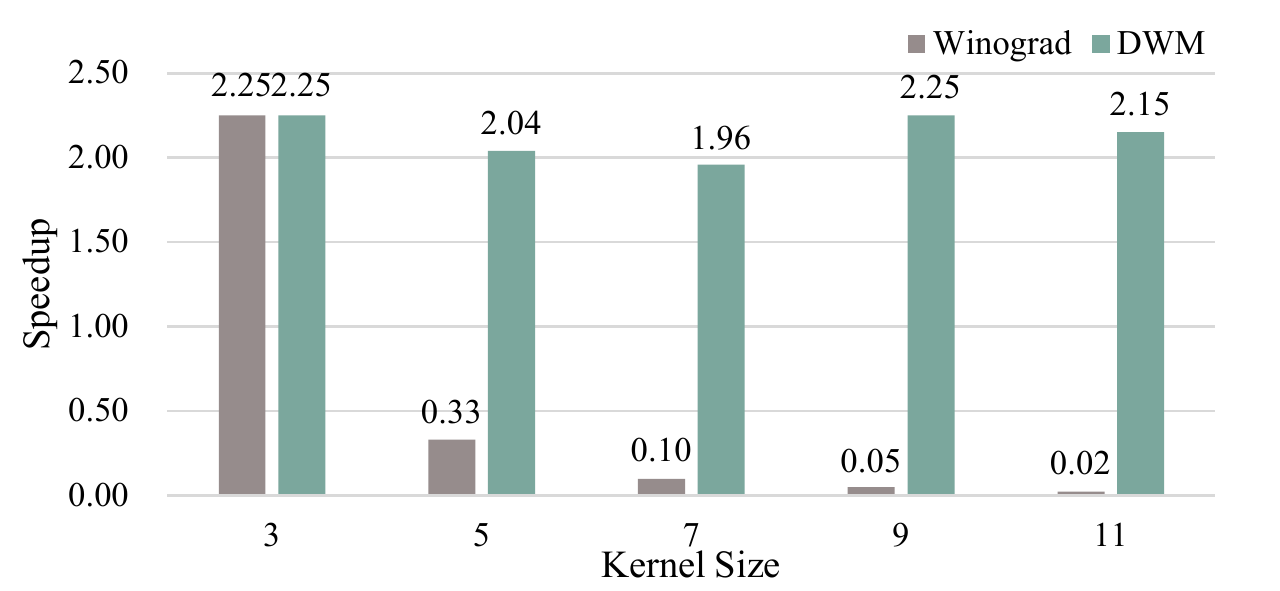}
        \caption{}\label{fig:1a}
    \end{subfigure}
    
    \begin{subfigure}{\columnwidth}
        \centering
        \includegraphics[width=0.95\columnwidth]{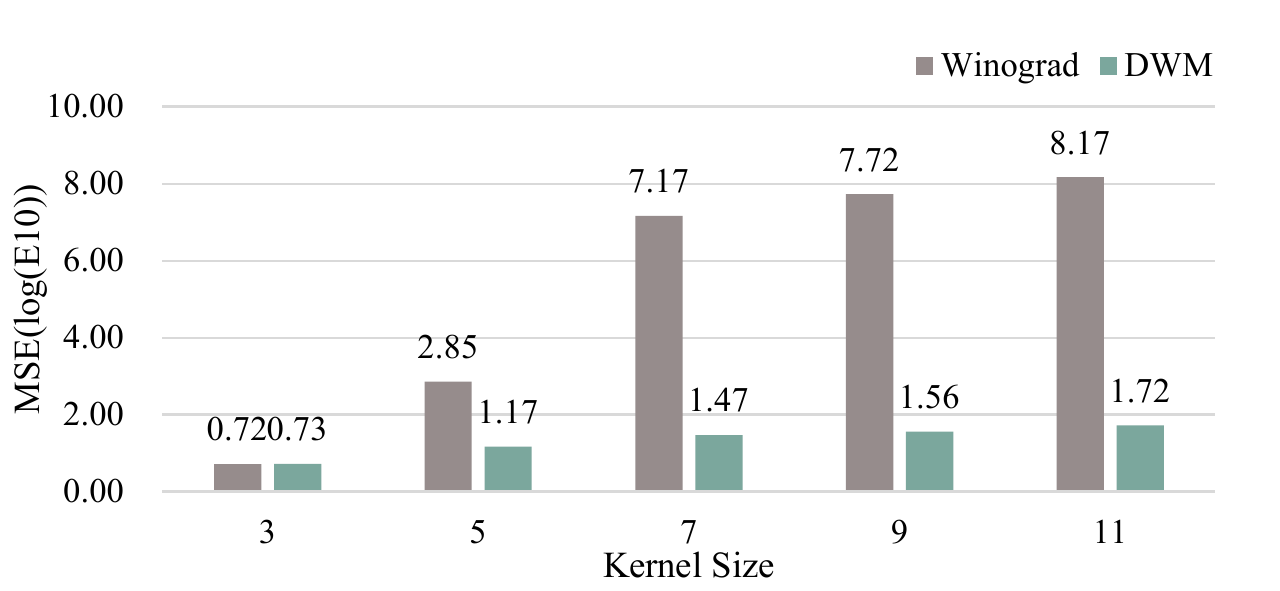}
        \caption{}\label{fig:1b}
    \end{subfigure}
    \caption{(a): The speedup comparison between two accelerating algorithms, measured by FLOPs. Baseline is the regular convolution result of FP64;  (b): The numerical accuracy of two accelerating algorithms, which is the mean squared error (MSE), scaled by $log_{10}(error) + 10$ to make it positive.}
    \label{fig:first}
\end{figure}

Lavin~\shortcite{lavin2016fast} applied Winograd's minimal filtering algorithm~\cite{winograd1980arithmetic} to half reduce the number of multiplications in convolutions. 
Unfortunately, Winograd's minimal filtering algorithm is only effective on $3 \times 3$ kernels with stride 1. 
When the kernels are larger than $3\times3$, the transform matrices of Winograd's minimal filtering algorithm will introduce much more decimal multiplications causing precision and efficiency problems.  
Another problem is that Winograd's minimal filtering algorithm cannot be used on convolutions with stride $>$ 1.
Since convolutions of kernel size larger than $3 \times 3$ and stride $>$ 1 are frequently used in CNNs, these two restrictions severely limited the application of Winograd's minimal filtering algorithm.

To tackle the above drawbacks, we propose the Decomposable Winograd Method (DWM) to extend the Winograd's minimal filtering algorithm into the cases of large kernels and stride $>$ 1. 
First, for kernels larger than 3, DWM decomposes the original kernel into several kernels smaller than 3, on which we can apply Winograd's minimal filtering algorithm separately.
Thus, for large kernels situation, DWM can still reduce the number of multiplications by 50\% and keep the numerical accuracy the same as the original convolution.
Second, for stride larger than 1 situation, DWM splits the kernels into several parts to apply Winograd's minimal filtering algorithm. 
\Eg, with stride 2, DWM is equal to split the polynomials of Winograd into odd ones and even ones and compute them respectively.
Therefore, DWM break through the kernel size and stride restriction of Winograd's minimal filtering algorithm.
DWM has the advantages of both computation and numerical accuracy.
DWM can efficiently reduce the multiplications of regular convolutions.
As shown in Figure~\ref{fig:first}(a), with the kernel size increasing from 3 to 11, the speedup of Winograd's minimal filtering algorithm decreases seriously, while the speedup of DWM keep to $\sim$ 2 .
Furthermore, DWM can keep the final result totally equivalent to the result of regular convolutions, which makes it suitable to be applied to actual products.
As shown in Figure~\ref{fig:first}(b), with the kernel size increasing from 3 to 11, the accuracy error of DWM keep small but the error of Winograd's minimal filtering algorithm increases quickly. 
These advantages of DWM enables the fast exploring of larger kernel size and larger stride value in CNNs for high performance and accuracy and even the potential for new CNNs.
Experiments show that DWM achieves $\sim2\times$ acceleration while keeping the numerical error under E-07, which is close to the numerical accuracy of FP32 convolution.

The contribution of this paper has three aspects:
\begin{itemize}
\item We identify the limitation of the Winograd's minimal filtering algorithm that it suffers from significantly increased FLOPs and numerical accuracy problem for kernel size larger than 3x3 and fails on convolution with stride larger than 1. 
\item We propose a novel DWM method which decomposes kernels with large size or large stride to several small kernels with stride as 1 for further apply Winograd method, so that DWM can reduce the number of multiplications while keeping the numerical accuracy.
\item We evaluate the proposed DWM method on convolutions with kernel size varying from 3 to 11 and stride from 1 to 2. Experimental results show that  DWM is able to support all kinds of convolutions with a speedup of $\sim$2, without affecting the numerical accuracy.
\end{itemize}
\section{Related Works}
So far, with the success of convolutional neuron networks, many researchers focus on accelerate convolution using linear algebra property. 
Cong and Xiao~\shortcite{cong2014minimizing} save 47\% amount of multiplications by utilizing the linear algebra property at the sub-matrix block level. 
Mathieu et al.~\shortcite{madisetti1997digital} first proposed the method of using Fast Fourier Transform (FFT) to reduce the computation of convolution operations.
After that, the FFT algorithm was refined by Vasilache~\shortcite{vasilache2014fast}.
Their two GPU implementations called cuFFT and fbfft outperformed the convolution library made by NVIDIA at that time.
Lavin~\shortcite{lavin2016fast} exploited several element-level linear algebra methods to reduce the number of multiplications, including Winograd's minimal filtering algorithm~\cite{winograd1980arithmetic} and FFT. 
Nowadays, cuDNN~\cite{chetlur2014cudnn}, a state-of-art deep learning library, includes both the Winograd algorithm and the FFT algorithm as their convolution implementation.

Some researchers made their efforts to overcome the defects of the Winograd algorithm. 
Some of them tried to overcome the incompatibility of the Winograd algorithm and model sparsification. 
Li~\shortcite{li2017enabling} proposed a method to sparse native Winograd coefficients and obtains sparsity level beyond 90\% with only 0.1\% accuracy loss. 
Liu~\shortcite{liu2018efficient} moved ReLU operation into the Winograd transformation and then pruned the weights in the Winograd domain. 
This approach reduced the number of multiplications by $10.8\times$ with loss of accuracy less than 0.1\%. 
Choi ~\shortcite{choi2018compression} proposed a prune principle to make the data keep sparse after the Winograd transformation. 
Some researchers attempted to solve the numerical accuracy and kernel size problem.
Barabasz~\shortcite{barabasz2019winograd} investigated a wider range of Winograd algorithms for DNNs, which significantly improve floating-point accuracy in many cases. 
In fp16, this approach has given up to 6.5 times better image recognition accuracy in one important case. 
Vincent~\shortcite{vincent2017improving} decreased the numerical error of large tile Winograd algorithm by selecting the polynomial points. 
Meng and Brothers~\shortcite{meng2019efficient} extended the Winograd algorithm to larger tiles by introducing complex numbers during the Winograd transformation.
Other researchers focused on the hardware implementation of the Winograd algorithm. 
Due to the low memory ceiling of GPU hardware, the Winograd algorithm can be used to speed up CPU convolution operations and achieves 5 to 25-fold improvement in throughput compared to previous state-of-art implementations~\cite{budden2017deep}. Besides, for the mobile CPU acceleration, the Winograd algorithm achieves up to 60\% performance improvements in the full network compared to im2col/im2row based optimization techniques~\cite{maji2019efficient}. 

Different from the researches mentioned above, our methods focus on the Winograd algorithm itself instead of the combination of the Winograd algorithm and other methods such as sparsification.
By extending the Winograd algorithm to a much wider situation, we efficiently reduce the number of multiplications while keeping the calculation's numerical accuracy stably high.
\section{Preliminary on the Winograd Algorithm}
\subsection{The Winograd Algorithm}
As an equivalent problem of multi-dimensional FIR filters problem, convolution can be implemented more efficiently using Winograd minimal filtering algorithm~\cite{lavin2016fast}.
Denoting the result of computing $m$ outputs with an $r$-tap FIR filter as $F(m, r)$, the corresponding convolution algorithm for it requires $m+r-1$ multiplications.

The original Winograd algorithm is derived from the relationship between polynomial multiplication and 1-D convolutions using the Chinese Remainder Theorem(CRT)~\cite{winograd1980arithmetic}. 
For the fixed $m$ and $r$, the whole algorithm contains three fixed transformation matrices: $A$, $B$ and $G$. 

Considering the original 1-D situation, the $r$ element filter $g(x)$ and $l=m+r-1$ element input signal $d(x)$ can be represented as polynomials ($0<r<l$):
{\small
\begin{equation}
\begin{aligned}
g(x)&=g_{r-1}x^{r-1}+g_{r-2}x^{r-2}+...+g_1x+g_0, \\
d(x)&=d_{l-1}x^{l-1}+d_{l-2}x^{l-2}+...+d_1x+d_0,
\label{eq:g_and_d}
\end{aligned}
\end{equation}
}
then the result of convolution $g(x)*d(x)$ can be obtained by calculating the coefficients of polynomial multiplication
{\small
\begin{equation}
\begin{aligned}
y(x)=g(x)d(x).
\end{aligned}
\end{equation}
}
Applying CRT, we can get three transformation matrices $A$, $B$ and $G$,
and the process of doing convolution can be formulated as the following:
{\small
\begin{equation}
Y=A^T[(Gg)\odot(B^Td)].
\end{equation}
}
where $Y$ denotes the convolution output and $\odot$ denotes element-wise multiplication. 
For 2-D convolutions, we can nest the F(m, r) with itself, and then get
{\small
\begin{equation}
\label{eq:Winograd}
Y=A^T[(GgG^T)\odot(B^TdB)]A.
\end{equation}
}
From equation~\eqref{eq:Winograd}, we can derive the gradient of neuron (denoted as $\nabla d$) and the gradient of weight (denoted as $\nabla g$) of Winograd algorithm using the chain rule:
{\small
\begin{equation}
\begin{aligned}
\label{eq:backward}
\nabla d&=B[(A\nabla YA^T)\odot(GgG^T)]B^T, \\
\nabla g&=G^T[(A\nabla YA^T)\odot(B^TdB)]G,
\end{aligned}
\end{equation}
}
where $\nabla Y$ is the gradient passed from the next layer. 
\subsection{Drawbacks}
\subsubsection{Large Kernel Size}
The benefit of the Winograd algorithm comes from the simplicity of transformation matrices. For example, applying the Winograd algorithm, the transformation matrices of $F(2, 3)$ are shown as follows:
{\small
\begin{equation}
\centering
\begin{aligned}
B^T&=        
\left[
    \begin{array}{rrrr}
      1 & 0 &-1 & 0 \\
      0 & 1 & 1 & 0 \\
      0 &-1 & 1 & 0 \\
      0 & 1 & 0 &-1
    \end{array}
\right], \\
G&=
\left[
    \begin{array}{rrr}
      1 & 0 & 0 \\
      1/2 & 1/2 & 1/2 \\ 
      1/2 & -1/2 & 1/2 \\
      0 & 0 & 1
    \end{array}
\right], \\
A^T&=
\left[
    \begin{array}{rrrr}
      1 & 1 & 1 & 0 \\
      0 & 1 &-1 &-1
    \end{array}
\right].
\end{aligned}
\end{equation}
}
However, considering $F(2, 5)$ as an example, the transformation matrices becomes something like
{\small
\begin{equation}
\centering
\begin{aligned}
B^T&=        
\left[
    \begin{array}{rrrrrr}
       4 & 0 & -5 & 0 & 1 & 0 \\
       0 & -4 & -4 & 1 & 1 & 0 \\
       0 & 4 & -4 & -1 & 1 & 0 \\
       0 & -2 & -1 & 2 & 1 & 0 \\
       0 & 2 & -1 & -2 & 1 & 0 \\
       0 & 4 & 0 & -5 & 0 & 1
    \end{array}
\right], \\
G&=
\left[
    \begin{array}{rrrrr}
       1/4 & 0 & 0 & 0 & 0 \\
       -1/6 & -1/6 & -1/6 & -1/6 & -1/6 \\
       -1/6 & 1/6 & -1/6 & 1/6 & -1/6 \\
       1/24 & 1/12 & 1/6 & 1/3 & 2/3 \\
       1/24 & -1/12 & 1/6 & -1/3 & 2/3 \\
       0 & 0 & 0 & 0 & 1 \\
    \end{array}
\right], \\
A^T&=
\left[
    \begin{array}{rrrrrr}
       1 & 1 & 1 & 1 & 1 & 0 \\
       0 & 1 & -1 & 2 & -2 & 1
    \end{array}
\right].
\end{aligned}
\end{equation}
}
The huge number of decimals in transformation matrices makes the Winograd transformation not only high consumption, but also less accurate. 
\subsubsection{Stride $>$ 1}
Another problem of the Winograd algorithm is that it cannot be applied to stride $>$ 1 convolutions, for it is derived from polynomial multiplication which indicates stride 1 convolutions. Therefore, although the Winograd algorithm can implement convolutions much more efficiently, it is always used on $3\times3$ and stride 1 convolutions only.
\section{The Decomposable Winograd Method}
In this section, we propose a series of techniques called the Decomposable Winograd Method(DWM) to apply the Winograd algorithm mentioned above on more general cases like larger kernels and stride $>$ 1.
\subsection{Forward}
\subsubsection{Large Kernel Size}
\begin{figure}[tb!]
    \centering
    \begin{subfigure}{0.85\columnwidth}
        \centering
        \includegraphics[width=\columnwidth]{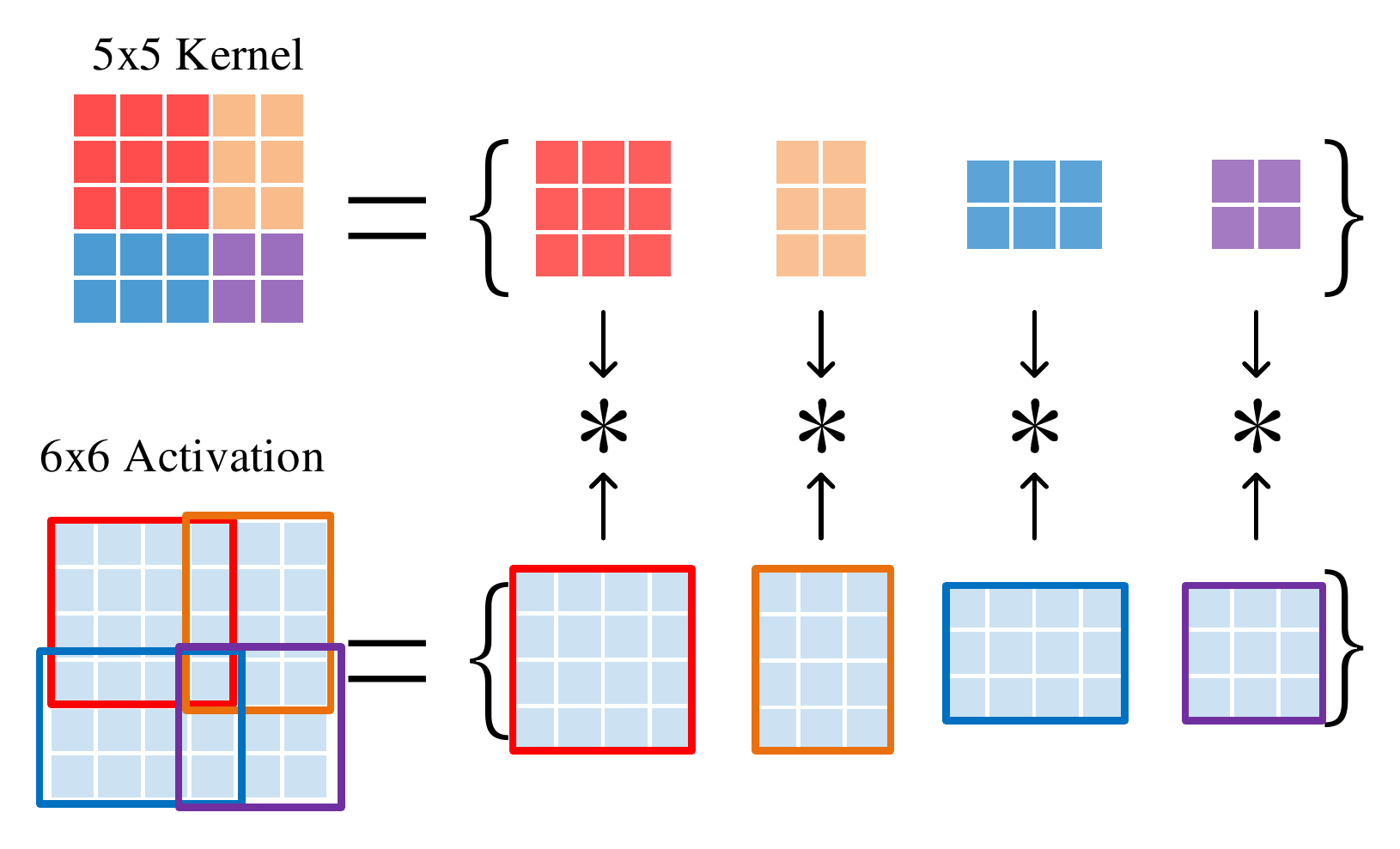}
        \caption{}\label{fig:method_large}
    \end{subfigure}
    
    \begin{subfigure}{\columnwidth}
        \centering
        \includegraphics[width=0.85\columnwidth]{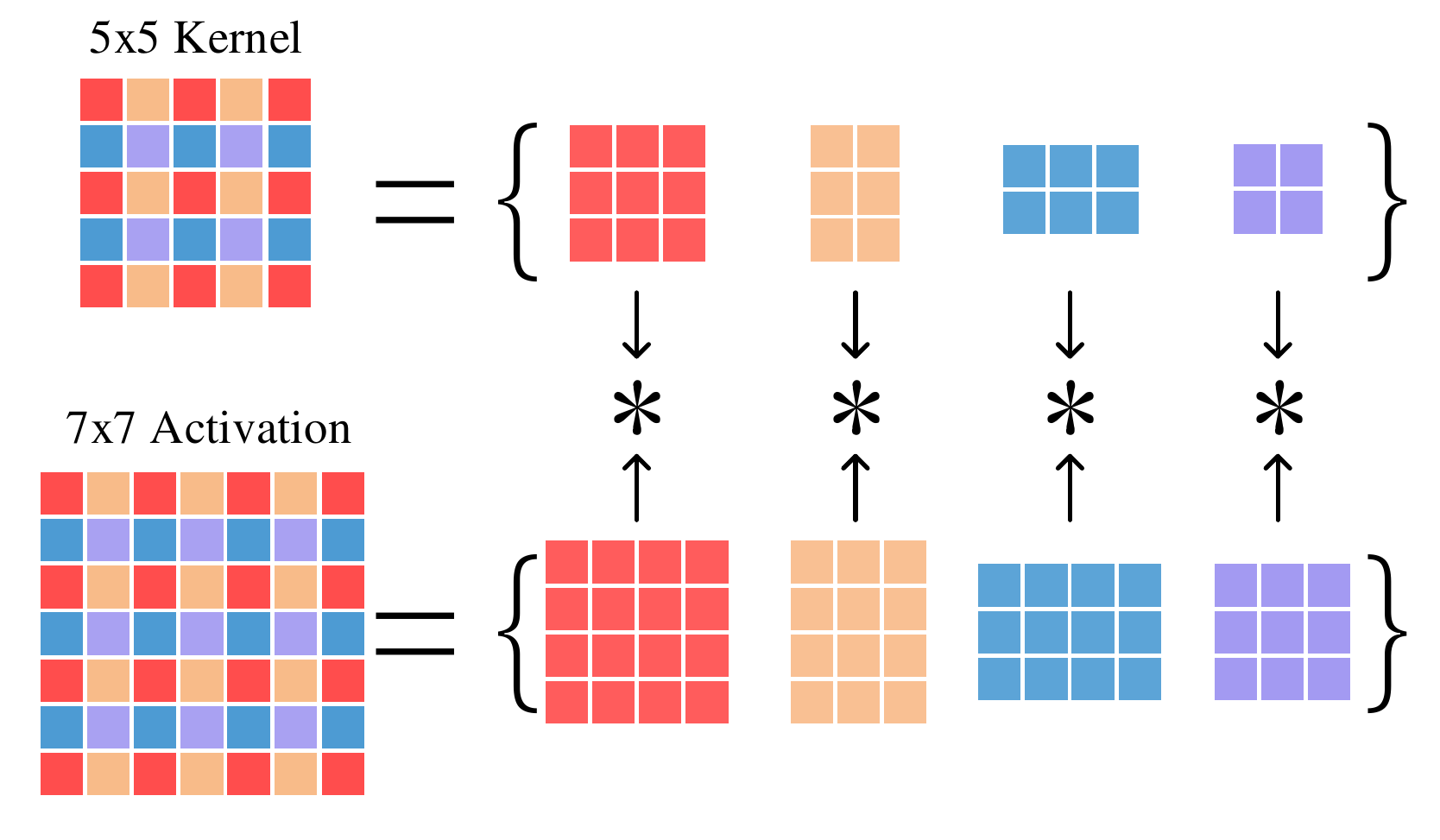}
        \caption{}\label{fig:method_stride}
    \end{subfigure}
    \caption{(a): Splitting a $5\times5$ and stride 1 convolution into four smaller convolutions; (b): Splitting a  $5\times5$ and stride 2 convolution into four stride 1 convolutions. '*' denotes doing convolution with corresponding parts.}
    \label{fig:method}
\end{figure}

\begin{figure}[tb!]
    \centering
    \includegraphics[width={0.9\columnwidth}]{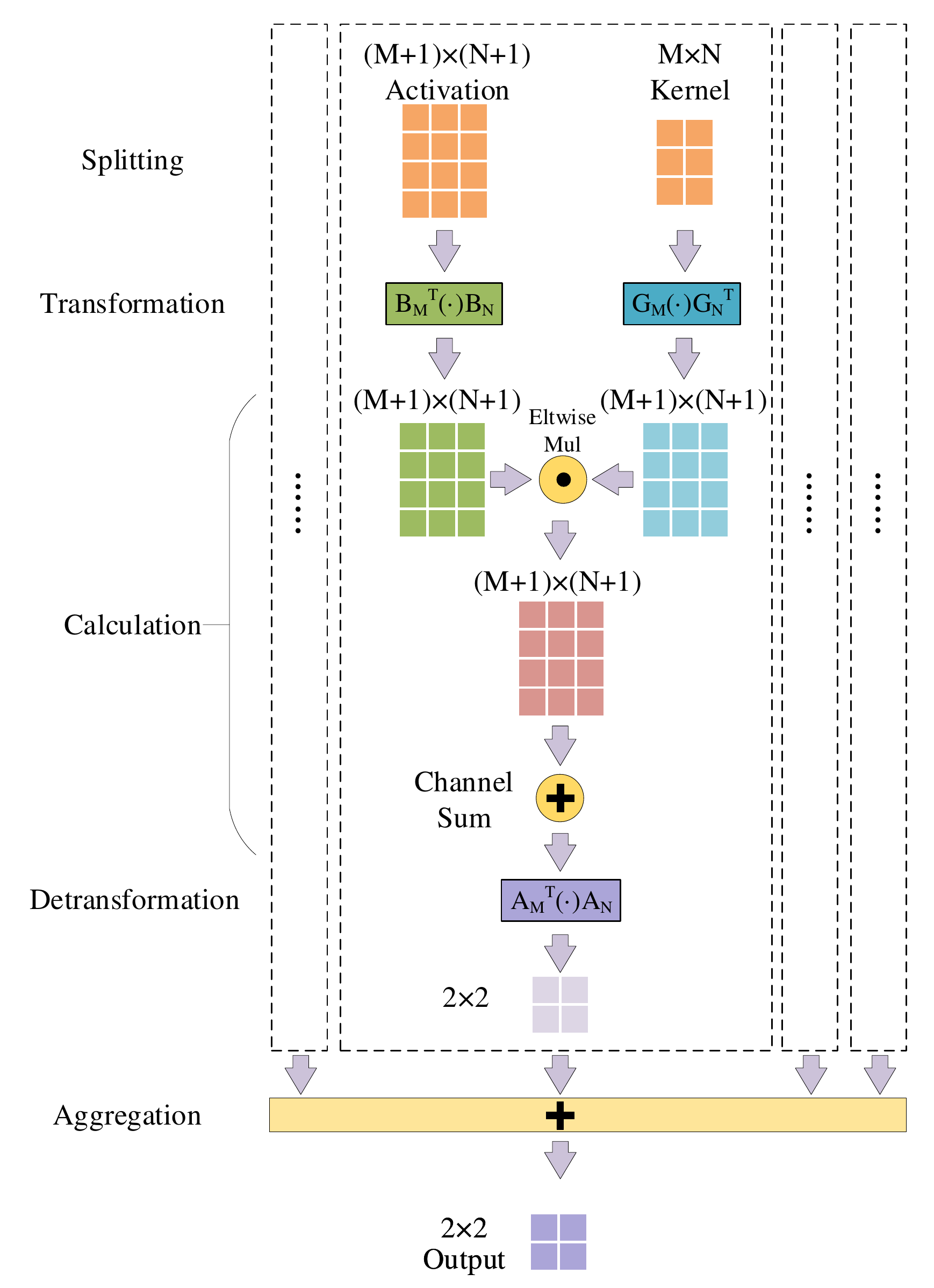}
    \caption{The process of doing convolution using Winograd algorithm. The four dotted frames denote convolution procedure of the four split parts, and each procedure can be divided into five steps: splitting, transformation, calculation, detransformation and aggregation. 
    }
    \label{fig:process}
\end{figure}

As we denote before, $g(x)$ represents the 1-D convolution filter and thus $r$ represents the size of convolution filter. Observing the derivation process above, we find that when $r$ becomes larger, $g(x)$ in equation~\eqref{eq:g_and_d} can be split into several small polynomials, to which we can apply the original Winograd algorithm:
{\small
\begin{equation}
\left\{
\begin{aligned}
    g^{(0)}(x)&=g_{2}x^2+g_{1}x+g_0 \\
    g^{(1)}(x)&=(g_{5}x^2+g_{4}x+g_3)x^3 \\
    \vdots \\
    g^{(\lfloor{r/3}\rfloor)}(x)&=\sum^{r-1\;mod\;3}_{i=0}g_
    {r-i-1}x^{(r-1\;mod\;3)-i} x^{3\lfloor{r/3}\rfloor}
\end{aligned}
\right..
\end{equation}
}
Then from $g(x)=\sum^{\lfloor{r/3}\rfloor}_{i=0} g^{(i)}(x) $ we can get
{\small
\begin{equation}
\begin{aligned}
y(x)&=g(x)d(x) 
=\sum^{\lfloor{r/3}\rfloor}_{i=0} [g^{(i)}(x)d(x)]
=\sum^{\lfloor{r/3}\rfloor}_{i=0} y^{i}(x),
\end{aligned}
\end{equation}
}
We can apply $F(2, 2)$ or $F(2, 3)$ on each $y^{i}(x)$ separately, with half multiplication reduced.
As for 2-D convolution, we can split the large kernel into small parts, and then apply Winograd algorithm to each part separately. The whole process is illustrated by Figure~\ref{fig:process}, which shows that we can process a common large kernel convolution in five steps: 
\begin{quote}   
\begin{itemize}
\item \textbf{Splitting.} Split the convolution kernel into several parts whose kernel size belows $3\times3$, and then prepare the input signal by slicing the redundant edges. This method is shown in Figure~\ref{fig:method_large}.
\item \textbf{Transformation.} Apply corresponding Winograd transformation $B^T(\cdot)B$ and $G(\cdot)G^T$ (in $F(2, 2)$ or $F(2, 3)$) on each part.
\item \textbf{Calculation.} Do element-wise multiplication and channel-wise summation.
\item \textbf{Detransformation.} Do the $A^T(\cdot)A$ inverse transformation to change the intermidiate results to spatial domain.
\item \textbf{Aggregation.} Sum the calculation results of each part, which gives the final result that equivalent to the original convolution. 
\end{itemize}
\end{quote}
For example, when $r=5$, we can split it into two parts
{\small
\begin{equation}
\left\{
\begin{aligned}
    g^{(0)}(x)&=g_{2}x^{2}+g_{1}x+g_0 \\
    g^{(1)}(x)&=(g_{4}x+g_{3})x^3 \\
\end{aligned}
\right., \\
\end{equation}
}
and $g(x)=g^{(0)}(x)+g^{(1)}(x)$.
Then we get:
{\small
\begin{equation}
y(x)=g(x)d(x)=g^{(0)}(x)d(x)+g^{(1)}(x)d(x).
\end{equation}
}
For $g^{(0)}(x)d(x)$, we can apply $F(2, 3)$ to it, and for $g^{(1)}(x)d(x)$, we can apply $F(2, 2)$.
When it comes to a 2-D $5\times5$ convolution case, we can split the kernel into 4 parts:$3\times3$, $3\times2$, $2\times3$ and $2\times2$, just as the method shown in Figure~\ref{fig:method_large}.
This method's advantage is that using $F(2, 2)$ and $F(2, 3)$ instead of larger ones not only reduces the multiplications of Winograd transformation efficiently but also keeps the Winograd algorithm's accuracy because of the few amounts of multiplications in transformation matrices. We will further illustrate this advantage in the experiment part.
\subsubsection{Stride $>$ 1}
Normal polynomial multiplication indicates stride 1 convolution, but we can surmount this barrier by grouping the polynomial into several parts. Denoting convolution stride as $s$, 1-D convolution kernel $g(x)$ with $r$ elements can be split into
{\small
\begin{equation}
\begin{aligned}
\left\{
\begin{aligned}
    g^{(0)}(x)&=\sum_{i=0}^{\lfloor (r-1)/s \rfloor} g_{s*i}x^{s*i} \\
    g^{(1)}(x)&=\sum_{i=0}^{\lfloor (r-2)/s \rfloor} g_{s*i+1}x^{s*i+1} \\
    \vdots \\
    g^{(s-1)}(x)&=\sum_{i=0}^{\lfloor (r-s-1)/s \rfloor} g_{s*i+s-1}x^{s*i+s-1}
\end{aligned}
\right..
\end{aligned}
\end{equation}
}
The input signal $d(x)$ with $l$ elements can be split into $d^{(0)}(x), d^{(1)}(x), ..., d^{(s-1)}(x)$similarly.

Then we get several stride 1 convolutions which can be represented by polynomials multiplications $y^{(i)}(x)=g^{(i)}(x)d^{(i)}(x)$ and $y(x)=\sum y^{(i)}(x)$. By applying Winograd algorithm to them, we have reduced the multiplications of 1-D convolutions with stride $>$ 1 successfully.
For 2-D convolutions, we nest the 1-D convolution methods.
This process also contains five parts: splitting, transformation, calculation, detransformation and aggregation, which is similar to Figure~\ref{fig:process}.
The splitting method is illustrated by Figure~\ref{fig:method_stride}.

For instance, when we are dealing with stride 2 convolutions, we can group the convolution kernel $g(x)$ and input signal $d(x)$ by their degree's parity, and then get
{\small
\begin{equation}
\begin{aligned}
\left\{
\begin{aligned}
    g^{(0)}(x)&=\sum_{i=0}^{\lfloor (r-1)/2 \rfloor} g_{2i}x^{2i} \\
    g^{(1)}(x)&=\sum_{i=0}^{\lfloor (r-2)/2 \rfloor} g_{2i+1}x^{2i+1}
\end{aligned}
\right.
\end{aligned}
\end{equation}
}
and 
{\small
\begin{equation}
\begin{aligned}
\left\{
\begin{aligned}
    d^{(0)}(x)&=\sum_{i=0}^{\lfloor (l-1)/2 \rfloor} d_{2i}x^{2i} \\
    d^{(1)}(x)&=\sum_{i=0}^{\lfloor (l-2)/2 \rfloor} d_{2i+1}x^{2i+1}
\end{aligned}
\right..
\end{aligned}
\end{equation}
}
In a 2-D case, a $5\times5$ stride 2 convolution on $7\times7$ activation can be splitted into doing four stride 1 convolutions: $3\times3$, $3\times2$, $2\times3$ and $2\times2$. 
Details are shown in Figure~\ref{fig:method_stride}.

Occasionally, we need to do a stride $>$ 2 convolution with large kernel size, then we can combine the two techniques mentioned above. By applying these two techniques, we can optimize all kinds of convolutions with the Winograd algorithm, reducing the number of multiplications by half.
\subsection{Backward}
Using equation~\eqref{eq:backward}, we can apply DWM in the process of calculating gradients, and this will also reduce the number of multiplications by half.
From SGD algorithm, we can derive that
{\small
\begin{equation}
\begin{aligned}
g^{wino}_{(t+1)}&=G(g_{(t)}+\nabla g_{(t)})G^T \\
&=Gg_{(t)}G^T+G(G^T[(A\nabla Y_{(t)}A^T)\odot(B^Td_{(t)}B)]G)G^T, \\
\end{aligned}
\end{equation}
}
where $W_{wino}$ denotes weight in Winograd form and $t$ denotes iteration. 
We should notice that $GG^T\neq E$, so we keep weight W in the normal form (or called 'spatial domain') instead of Winograd form (or called 'frequency domain') during the training process in order to make DWM be the equivalent substitute of normal convolution.
\begin{table*}[t]
\caption{Mean squared error (MSE) between different acceleration algorithms and the FP64 result by running a forward convolution, i.e FP64 convolution result is the baseline. H/W means the size of featur map, and FP32/FP16 indicates doing convolution in FP32/FP16 format.}\smallskip
\centering
\resizebox{.95\textwidth}{!}{
\smallskip\begin{tabular}{c|c|c|c|c|c|c|c|c|c}
\toprule
Kernel Size & H/W & Channel & Filters & FP32 & FP16 & Winograd FP32 & Winograd FP16 & DWM FP32 & DWM FP16\\
\hline
3x3   & 14 & 256 & 256 & 2.21E-08 & 2.71E-04 & 5.24E-10 & 1.44E-03 & 5.32E-10 & 3.42E-02 \\
3x3   & 28 & 128 & 128 & 1.11E-10 & 1.41E-04 & 1.43E-10 & 7.48E-04 & 1.47E-10 & 9.08E-03 \\
5x5   & 14 & 256 & 256 & 1.05E-02 & 6.93E-04 & 7.14E-08 & 1.07E-01 & 1.47E-09 & 9.72E-02 \\
5x5   & 28 & 128 & 128 & 3.15E-10 & 3.80E-04 & 2.00E-08 & 5.78E-02 & 4.33E-10 & 2.83E-02 \\
7x7   & 14 & 256 & 256 & 6.13E-10 & 1.25E-03 & 1.47E-03 & NaN      & 2.97E-09 & 1.97E-01 \\
7x7   & 28 & 128 & 128 & 5.61E-10 & 7.16E-04 & 4.24E-04 & NaN      & 8.86E-10 & 5.88E-02 \\
9x9   & 14 & 256 & 256 & 9.90E-10 & 1.90E-03 & 5.31E-03 & NaN      & 3.67E-09 & 2.36E-01 \\
9x9   & 28 & 128 & 128 & 8.52E-10 & 1.14E-03 & 1.62E-03 & NaN      & 1.18E-09 & 7.33E-02 \\
11x11 & 14 & 256 & 256 & 1.47E-09 & 2.60E-03 & 1.48E-02 & NaN      & 5.30E-09 & 3.46E-01 \\
11x11 & 28 & 128 & 128 & 1.15E-09 & 1.63E-03 & 4.35E-03 & NaN      & 1.81E-09 & 1.15E-01 \\
\bottomrule
\end{tabular}
}
\label{table:per_layer_acc}
\end{table*}
Considering DWM, the two techniques mentioned above, we can derive the corresponding back propagation rules. Denoting output as $Y$, the two techniques have the same form in the aggregation step:
{\small
\begin{equation}
\begin{aligned}
Y = \sum_i Y^{(i)}.
\end{aligned}
\end{equation}
}
From the derivative rules we know that for part $j$:
{\small
\begin{equation}
\begin{aligned}
\nabla Y^{(j)} &= \nabla^{(j)} Y^{(j)} + \nabla^{(j)}\sum_{i\neq j} Y^{(i)} \\
&=\nabla^{(j)}\sum_i Y^{(i)} 
=\nabla^{(j)} Y.
\end{aligned}
\end{equation}
}
This means that we don't need to store each output $Y^{(i)}$ for the backward.
Furthermore, it is easy to derive that in the splitting step, the backward acts as follows: 
{\small
\begin{equation}
\begin{aligned}
\nabla d=\sum \nabla d^{(i)}, 
\nabla g=\sum \nabla g^{(i)},
\end{aligned}
\end{equation}
}
where i denotes different parts produced by DWM.

\subsection{Comparison and Discussion}
We are not the first one who tries to solve the large kernel size problem that appeared during applying the Winograd algorithm. 
Lu et al.~\cite{lu2017evaluating} tried to implement the Winograd algorithm on FPGAs, and they solved large kernel problems by padding the kernel. 
However, the padded elements will be filled with non-zero values after the Winograd transformation, which means that paddings will bring lots of extra calculations.
For DWM, we find a way that precisely separates the convolution operations without padding. This method avoids any redundant floating-point multiplications during Winograd transformation, and thus achieves the best acceleration without any numerical accuracy loss.
\begin{table*}[t]\small
\caption{Top-1 accuracy, FLOPs and speedup of several acclerating algorithm on different networks. Origin means the original top-1 accuracy and FLOPs of networks.}\smallskip
\centering
\smallskip\begin{tabular}{c|c c|c c c|c c c}
\toprule
\multirow{2}{*}{Network} & \multicolumn{2}{|c}{Origin} & \multicolumn{3}{|c}{Winograd} & \multicolumn{3}{|c}{DWM} \\
                & Acc & GFLOPs & Acc & GFLOPs           & speedup & Acc & GFLOPs & speedup    \\
\hline
AlexNet~\cite{krizhevsky2012imagenet}         & 56.52 & 0.71  & 56.51 & 0.56 & 1.28 & 56.51 & 0.45 & 1.57 \\
GoogLeNet~\cite{szegedy2015going}       & 69.79 & 1.51  & 69.79 & 0.97 & 1.55 & 69.77 & 0.92 & 1.65 \\
Inception-V3~\cite{szegedy2016rethinking}    & 69.54 & 2.86  & 69.47 & 2.34 & 1.22 & 69.46 & 1.92 & 1.49 \\
ResNet-152~\cite{he2016deep}      & 78.31 & 11.62 & 78.31 & 8.78 & 1.32 & 78.31 & 8.60 & 1.35 \\
DenseNet-161~\cite{huang2017densely}    & 77.14 & 7.88  & 77.13 & 6.32 & 1.25 & 77.12 & 6.23 & 1.26 \\
ProxylessGPU~\cite{cai2018proxylessnas}    & 75.08 & 0.49  & 74.77 & 0.49 & 1.01 & 75.06 & 0.47 & 1.05 \\
ProxylessMobile~\cite{cai2018proxylessnas} & 74.59 & 0.35  & 74.47 & 0.34 & 1.01 & 74.57 & 0.33 & 1.06 \\
\bottomrule
\end{tabular}
\label{table:network_analysis}
\end{table*}

Furthermore, with the rise of neural architecture search (NAS), convolutions with larger kernel sizes such as $5\times5$ or $7\times7$ have become more popular than ever. Networks like ProxylessNAS~\cite{cai2018proxylessnas} show that different computation environments will breed different neural architectures. 
For example, when we want to implement our neural networks on GPUs rather than on mobile CPUs, large kernel sizes may be more suitable choices than small ones because of the computation parallelism of GPUs. 
From beginning to the end, the architecture of neural networks follows the most popular hardware, not the other way round. 
Hence, we believe that reducing the computation overhead of neural networks is worthy in any case. 
\section{Experiments}
\subsection{Setup}
All the results were tested on NVIDIA V100 GPU. 
We have implemented DWM on TensorFlow~\cite{abadi2016tensorflow} and PyTorch~\cite{paszke2017automatic}. 
On both platforms, the implemented DWM performs like a plug-and-play operator, which makes it convenient to use during inference and training. 
We tested the numerical accuracy of the algorithms by doing convolution with the standard normal distribution random numbers on one single layer, and then calculate the mean squared error (MSE) with FP64 convolution results. 
The numerical accuracy on a single layer was tested on both two platforms, and the results were the same.
For all single layer tests, the batch size is set to 256 and the layers are the same padded. 
We measured the accuracy and FLOPs of networks based on PyTorch. 
The traditional model architectures are gained from TorchVision, and the ProxylessNAS~\cite{cai2018proxylessnas} analysis is based on the official PyTorch implementation.
The Networks' accuracy was measured on ImageNet 2012~\cite{ILSVRC15}.
\subsection{Numerical Accuracy of Single Layer}
We estimated the mean squared error (MSE) between several methods and the FP64 results by doing a forward convolution. 
The input signal and convolution weights are random numbers generated by standard normal distribution using Numpy, set seed 11. 
We assumed two application situations: larger feature map (set $28\times28$) with fewer channels (set $128$) and smaller feature map (set $14\times14$) with more channels (set $256$), which is consistent with the reality. 
As shown in Table~\ref{table:per_layer_acc}, 
(1) DWM has better numerical accuracy than the traditional Winograd algorithm almost in all situations.
It is obvious that the traditional Winograd algorithm faced with a serious numerical accuracy problem as the kernel size grows up. 
When the kernel size is $7\times7$ or larger, the error of the FP32 Winograd algorithm approaches FP16's, which may cause accuracy problems.
By contrast, DWM's numerical error stays at a low level, which is close to the result of FP32, meaning that it can be applied to all kinds of convolution operations without any problem.
(2) When using FP16, the traditional Winograd algorithm may get an overflow.
This may be caused by the intermediate results of the Winograd transformation.
When kernel size becomes larger, the transformation matrices will be filled with large numbers, and these numbers may make the results of matrix multiplication become too large to be represented in FP16. 
This also shows the advantages of using DWM instead of the traditional Winograd algorithm.
Furthermore, the FP16 DWM is implemented without any optimization, which means the numerical accuracy of it still can be improved.
(3) In some situations, DWM FP32 is more accurate than FP32 ones.
This is reasonable because DWM consumes fewer multiplications, which may make it has better numerical accuracy.
\subsection{FLOPs Estimation on Single Layer}
\begin{table*}[hbtp]
\caption{The speedup of several acclerating algorithms on different kinds of convolutions. We assume that the output size is fixed to 14$\times$14.} \smallskip
\centering
\resizebox{.6\textwidth}{!}{
\smallskip\begin{tabular}{c|c|c|c c|c c}
\toprule
\multirow{2}{*}{Kernel Size} & \multirow{2}{*}{Stride} & Direct & \multicolumn{2}{|c}{Winograd} & \multicolumn{2}{|c}{DWM} \\
      &     & FLOPs   &FLOPs      &speedup&FLOPs&speedup \\
\hline
3x3   & 1 & 1.76E+03 & 784      & 2.25 & 784      & 2.25 \\
5x5   & 1 & 4.90E+03 & 1.48E+04 & 0.33 & 2.40E+03 & 2.04 \\
7x7   & 1 & 9.60E+03 & 9.72E+04 & 0.10 & 4.90E+03 & 1.96 \\
9x9   & 1 & 1.59E+04 & 3.16E+05 & 0.05 & 7.06E+03 & 2.25 \\
11x11 & 1 & 2.37E+04 & 1.07E+06 & 0.02 & 1.10E+04 & 2.15 \\
3x3   & 2 & 1.76E+03 & N/A      & N/A  & 1.23E+03 & 1.44 \\
5x5   & 2 & 4.90E+03 & N/A      & N/A  & 2.40E+03 & 2.04 \\
7x7   & 2 & 9.60E+03 & N/A      & N/A  & 4.90E+03 & 1.96 \\
9x9   & 2 & 1.59E+04 & N/A      & N/A  & 8.28E+03 & 1.92 \\
11x11 & 2 & 2.37E+04 & N/A      & N/A  & 1.10E+04 & 2.15 \\ 
\bottomrule
\end{tabular}
}
\label{table:per_layer_flops}
\end{table*}
\begin{table*}[htbp]
\caption{The actual runtime of several acclerating algorithms on different kinds of convolutions tested by nvprof. The batch size, the channels and the filters are 256. The input size is fixed to 14$\times$14.} \smallskip
\centering
\resizebox{.7\textwidth}{!}{
\smallskip\begin{tabular}{c|c|c|c|c|c}
\toprule
Kernel Size & DWM(ms) & Winograd(ms) & cuDNN(ms) & Wino/DWM & cuDNN/DWM \\
\hline
3x3         & 3.67    & 3.35         & 2.80      & 0.91     & 0.76      \\
5x5         & 11.37   & 69.70        & 11.26     & 6.13     & 0.99      \\
7x7         & 22.83   & 133.34       & 24.51     & 5.84     & 1.07      \\
9x9         & 30.67   & 248.71       & 50.92     & 8.11     & 1.66      \\
11x11       & 48.29   & 349.33       & 94.63     & 7.23     & 1.96      \\
\bottomrule
\end{tabular}
}
\label{table:per_layer_speed}
\end{table*}
We calculated the FLOPs of convolutions with different kernel sizes and 2 kinds of stride.
Due to the decimals in the transformation matrix in the traditional Winograd algorithm, the FLOPs caused by transformation cannot be ignored, and thus we only eliminate the influence of $0, \pm1, \pm2^n$ and $\pm\frac{1}{2^n}$ which can be easily implemented by shifting.
As shown in Table~\ref{table:per_layer_flops}, we can get the following results:
(1) DWM cost less computation than the traditional Winograd algorithm in all situations.
As the kernel size becomes larger, the FLOPs of traditional Winograd algorithm increases heavily, and most FLOPs concentrates on the Winograd transformation because the transformation process becomes a non-sparsity matrix multiplication.
On the contrary, the speedup of DWM keeps steady for its simple transformation matrices.
(2) DWM speeds up stride 2 convolutions by 2 $\times$, which cannot be achieved by the traditional method.
Not surprisingly, due to the splitting method of DWM, the speedup of stride 2 convolution still holds stably.
These advantages lead to stably speedup on almost all kinds of convolutions.

We also tested the actual runtime of convolution operations based on naive implementations of different kinds of convolutions. 
The result was tested by nvprof, an NVIDIA profiling tool. According to Table~\ref{table:per_layer_speed}, we can conclude that: 
(1) DWM outperforms the original Winograd algorithm, especially in large kernel situation. 
(2) DWM is faster than cuDNN in some situations. Actually, when the size of the feature map increases to around 100, cuDNN performs better than DWM (not presented due to the space left). 
The instability of cuDNN may be caused by some other accelerating algorithms. Hence, both DWM and cuDNN have their advantages.
Furthermore, the naive DWM implementation can be optimized by kernel fusion, soft pipeline and some other optimization techniques unlike the simple traditional 3x3 Winograd algorithm. Since we mainly focuses on speed up Winograd through the algorithm aspect in this paper, the implementation optimization of DWM is our future work.
\subsection{Total Analysis on Networks}
Finally, we analyzed several representative networks.
This analysis includes the comparison of total FLOPs of the network and the top-1 accuracy of inference on the ImageNet 2012 dataset.  
The top-1 accuracy of the two accelerating algorithms is tested using FP16.
As shown in Table~\ref{table:network_analysis}, we can conclude that:
(1) the top-1 acc is very close.
This result is not surprising, because Table~\ref{table:per_layer_acc} shows that only large kernel size will make DWM and the traditional Winograd algorithm different. Thus, the networks most of which consist of $3\times3$ kernel convolutions like will get similar inference results.
(2) When there are more large kernels in the network such as $11\times11$ in AlexNet and $1\times5/7$ or $5/7\times1$ in GoogLeNet and Inception-v3, DWM performs better. 

However, in some cases such as ResNet-152, most of the computation is produced by $3\times3$ kernels.
Worse, when it comes to some modern NAS architectures, the calculation is concentrated on $1\times1$ kernels because of the separable convolution. 
Over 90\% amount of calculation of convolutional neural networks is caused by convolution operations.
Hence, most of the architectures are designed or searched based on separable convolutions to reduce the amouont of calculation.
Although separable convolutions can cut the FLOPs down effectively, it reduces the representation of the original neural networks.
When the computing power grows up and more fast convolution algorithms are invented, FLOPs will not be the main consideration of architecture designing.


\section{Conclusion}
In this paper, we propose a novel DWM to extend Winograd's minimal filtering algorithm to a wide and general convolutions. 
Winograd's minimal filtering algorithm has been widely used to reduce the number of multiplications for faster processing. 
However, it has the drawbacks of sufferring from significantly increased FLOPs and numerical accuracy problem for kernel size larger than 3x3 and failing on convolution with stride larger than 1, so it is only effective on convolutions with kernel size as 3x3 and stride as 1.
To solve this problems, we propose DWM to break through the limitation of original Winograd's minimal filtering algorithm on convolutions of large kernel and large stride. 
DWM decomposes kernels with large size or large stride to several small kernels with stride as 1 for further applying Winograd method, so that DWM can reduce the number of multiplications while keeping the numerical accuracy.
Experimental results show that the proposed DWM is able to support all kinks of convolutions with a speedup of $\sim$2, without affecting the numerical accuracy.
These good properties of DWM enables the fast exploring of larger kernel size and larger stride value in CNNs for high performance, accuracy and even the potential for new CNNs.
\section{Acknowledgments}
This work is partially supported by the National Key Research and Development Program of China (under Grant 2017YFA0700902), the NSF of China (under Grants 61432016, 61532016, 61672491, 61602441, 61602446, 61732002, 61702478, 61732007, 61732020 and 61906179), Beijing Natural Science Foundation (JQ18013), the 973 Program of China (under Grant 2015CB358800), National Science and Technology Major Project (2018ZX01031102), the Transformation and Transfer of Scientific and Technological Achievements of Chinese Academy of Sciences (KFJ-HGZX-013), Key Research Projects in Frontier Science of Chinese Academy of Sciences (QYZDB-SSW-JSC001) , Strategic Priority Research Program of Chinese Academy of Science (XDB32050200, XDC01020000) and Standardization Research Project of Chinese Academy of Sciences (BZ201800001).
\bibliographystyle{aaai}
\bibliography{bibfile}
\end{document}